\documentclass{article}
\usepackage{arxiv}
\makeatletter\renewenvironment{table}{\@float{table}}{\end@float}\makeatother


\usepackage[utf8]{inputenc}
\usepackage[T1]{fontenc}
\usepackage[dvipsnames]{xcolor}
\usepackage{graphicx}
\usepackage{amsmath,amssymb,booktabs,xspace}
\usepackage[numbers,sort&compress]{natbib}
\usepackage[format=plain,labelformat=simple,labelsep=period,font=small,compatibility=false]{caption}
\usepackage[font=footnotesize,skip=3pt,subrefformat=parens]{subcaption}
\usepackage[hyphens]{url}
\usepackage{adjustbox}
\usepackage[shortlabels,inline]{enumitem}
\setlist[itemize]{noitemsep,leftmargin=*,topsep=0em}
\setlist[enumerate]{noitemsep,leftmargin=*,topsep=0em}

\usepackage{microtype}

\usepackage{amsmath,amssymb,amsthm,booktabs,multirow,xspace}
\usepackage{colortbl}

\newcommand{\parahead}[1]{\par\noindent\textbf{#1}\space\ignorespaces}

\usepackage{pifont}

\newcommand{\hstackl}[2]{\begin{tabular}[b]{@{}l@{}}#1\\#2\end{tabular}}
\newcommand{\hstackc}[2]{\begin{tabular}[b]{@{}c@{}}#1\\#2\end{tabular}}
\newcommand{\wallBasePeak}{0.56}
\newcommand{\wallBasePeakIt}{11}
\newcommand{\cubeBasePeak}{0.68}
\newcommand{\twoBasePeak}{0.80}
\newcommand{\wallEarly}{0.56}
\newcommand{\cubeEarly}{0.68}
\newcommand{\twoEarly}{0.77}
\newcommand{\wallBase}{0.39}
\newcommand{\wallOurs}{0.76}
\newcommand{\cubeBase}{0.64}
\newcommand{\cubeOurs}{0.79}
\newcommand{\twoBase}{0.77}
\newcommand{\wallFixed}{0.84}
\newcommand{\clCubeBase}{0.70}
\newcommand{\clCubeOurs}{0.81}
\newcommand{\wallFixedAll}{0.82}
\newcommand{\wallFixedImg}{0.85}
\newcommand{\ovhWall}{0.2}
\newcommand{\ovhCube}{3.1}
\newcommand{\idmShareCube}{4.3}
\newcommand{\idmTrainWall}{4.5}
\newcommand{\idmTrainCube}{15}
\newcommand{\bnCemOpen}{0.76}
\newcommand{\bnCheckOpen}{0.85}
\newcommand{\bnLossReleased}{0.066}
\newcommand{\bnLossRecal}{0.012}

\newcommand{\relWallBase}{0.88}
\newcommand{\relWallAcid}{0.76}
\newcommand{\relWallOurs}{0.89}
\makeatletter
\DeclareRobustCommand\onedot{\futurelet\@let@token\@onedot}
\def\@onedot{\ifx\@let@token.\else.\null\fi\xspace}
\def\etal{\emph{et al}\onedot}
\makeatother

\definecolor{linkblue}{rgb}{0.21,0.49,0.74}
\usepackage[pagebackref,breaklinks,colorlinks,allcolors=linkblue]{hyperref}
\usepackage[capitalize]{cleveref}
\crefname{section}{Sec.}{Secs.}
\Crefname{section}{Section}{Sections}
\Crefname{table}{Table}{Tables}
\crefname{table}{Tab.}{Tabs.}

\title{A Plan or an Adversarial Example?\\A Round-Trip Check for Latent World Models}
\renewcommand{\shorttitle}{A Plan or an Adversarial Example?}
\author{
  Huiyuan Tian$^{1}$ \quad Chih-Yung WEN$^{2}$ \quad Shijian Li$^{1,*}$ \\[4pt]
  $^{1}$College of Computer Science and Technology, Zhejiang University \\
  $^{2}$Department of Aeronautical and Aviation Engineering, The Hong Kong Polytechnic University \\[2pt]
  \texttt{tianhuiyuan@zju.edu.cn}, \quad \texttt{cywen@polyu.edu.hk}, \quad \texttt{shijianli@zju.edu.cn} \\[2pt]
  $^{*}$Corresponding author
}
\date{}

\begin{document}
\maketitle
\begin{abstract}
If we optimize an image until a classifier is confident it shows a cat, we may get an adversarial example, not a cat.
Planning with a latent world model is a similar search for actions whose predicted future is closest to the goal.
Is the result a plan, or an adversarial example of the world model?
We show that it can be either.
On the released DINO-WM and LeWM, more iterations of the cross-entropy method (CEM) lower the predicted cost but lose up to half of the problems CEM had already solved.
The model's own predictions do not reveal these failures.
We propose a simple test that executes the plan after every iteration and counts the problems a planner solves and then loses.
Unlike an image, a plan comes with its actions, which our round-trip check reads back from the predicted states with a small network, penalizing mismatches.
The check reads back the sum of each block of five actions, the part that two consecutive predicted states determine.
Without retraining the world model and at only \ovhWall\% to \ovhCube\% more planning time, it improves the widely used planners CEM and MPPI on three released world models, and gains more where planners lose more problems.
Without replanning, success rises from \wallBase{} to \wallOurs{} on Wall with DINO-WM and from 0.56 to 0.94 for MPPI on JEPA-WM.
With replanning (MPC), it rises from \clCubeBase{} to \clCubeOurs{} on Cube with LeWM.%
\end{abstract}

\begin{center}
  \captionsetup{type=figure}
  \begin{subfigure}[b]{0.2382\linewidth}
    \centering
    \includegraphics[width=\linewidth]{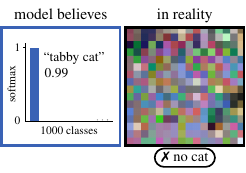}
    \caption{An adversarial example}
    \label{fig:banner:a}
  \end{subfigure}\hfill
  \begin{subfigure}[b]{0.3585\linewidth}
    \centering
    \includegraphics[width=\linewidth]{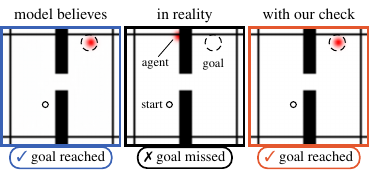}
    \caption{A plan on Wall with DINO-WM}
    \label{fig:banner:b}
  \end{subfigure}\hfill
  \begin{subfigure}[b]{0.3585\linewidth}
    \centering
    \includegraphics[width=\linewidth]{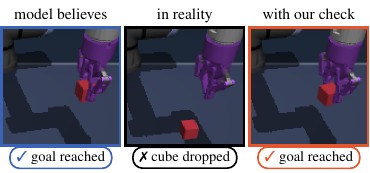}
    \caption{A plan on Cube with LeWM}
    \label{fig:banner:c}
  \end{subfigure}
  \caption{\textbf{A plan can be an adversarial example of the world model. The model cannot recognize it, and our round-trip check repairs the planner.}
  (a) A mosaic of color blocks optimized against an ImageNet~\cite{deng2009imagenet} classifier (ViT-B/16~\cite{dosovitskiy2021vit}) is labeled a tabby cat with probability 0.99.
  (b) The standard CEM planner optimizes actions against DINO-WM~\cite{zhou2025dinowm}.
  Its plan reaches the goal after 10 iterations and becomes an adversarial example after 30.
  The model predicts that the final plan reaches the goal, but in the simulator it ends on the wrong side of the wall.
  (c) The same happens with LeWM~\cite{lewm2026}, whose final plan drops the cube far from its target.
  With our round-trip check, the same planner's final plans reach the goal, without retraining the world model and at only \ovhWall\% and \ovhCube\% more planning time.}
  \label{fig:banner}
\end{center}

\section{Introduction}
\label{sec:intro}

\begin{figure}[t]
  \centering
    \begin{subfigure}[t]{0.485\linewidth}
      \centering
      \includegraphics[width=\linewidth]{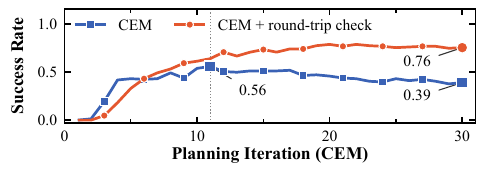}
      \caption{Success rate of the plan after each planning iteration}
      \label{fig:teaser:a}
    \end{subfigure}\hfill
    \begin{subfigure}[t]{0.485\linewidth}
      \centering
      \includegraphics[width=\linewidth]{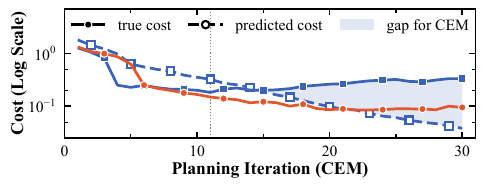}
      \caption{Predicted (dashed) and true (solid) cost of the same plan}
      \label{fig:teaser:b}
    \end{subfigure}%
  \caption{\textbf{With the standard planner, planning longer makes the model more confident and the plan worse.}
  The data are from DINO-WM on the Wall task with 148 problems.
  (a) With the standard CEM planner, the success rate peaks at \wallBasePeak{} after \wallBasePeakIt{} iterations and falls to \wallBase{} after 30, and with our round-trip check it rises to 0.76.
  (b) The median cost that the model predicts for the plan of CEM keeps falling, while the cost that the plan achieves in the simulator rises after iteration 11, and the shaded gap between the two widens.
  With the round-trip check, the true cost ends at less than a third of that of CEM.
  The dotted line marks iteration 11 in both panels.}
  \label{fig:teaser}
\end{figure}

As \Cref{fig:banner:a} shows, if we optimize an image until a classifier is confident that it shows a cat, the image need not look like a cat at all~\cite{nguyen2015fooled}.
Such images are adversarial examples in a broad sense~\cite{szegedy2014intriguing,goodfellow2015explaining}.
Planning with a latent world model~\cite{zhou2025dinowm,assran2025vjepa2,sobal2025pldm} is a similar optimization.
The world model predicts the future of a sequence of actions in the feature space of a visual encoder.
A planner such as the cross-entropy method (CEM)~\cite{rubinstein1999cem} searches for the sequence whose predicted future is closest to a goal image.
The actions correspond to the image, the goal to the class, and the cost that the model predicts to the classifier's score.
Is the result a plan, or an adversarial example of the world model?

As \Cref{fig:banner:b,fig:banner:c} illustrate, it can be an adversarial example.
We call a plan \emph{adversarial} if the model predicts that it reaches the goal and its execution does not, and \emph{genuine} if its execution does.
We find that a longer search can turn a genuine plan into an adversarial one.
As \Cref{fig:teaser} shows, on the Wall task of DINO-WM~\cite{zhou2025dinowm} the success of CEM's plan peaks at \wallBasePeak{} after \wallBasePeakIt{} iterations and falls to \wallBase{} after 30, while the cost that the model predicts keeps falling.
We find the same divergence on Cube and TwoRoom with LeWM~\cite{lewm2026}.
In seven of the thirteen combinations of world model, task and planner that we test, the final plan fails 14\% to 68\% of the problems an earlier plan had solved.
These seven span three released world models and two widely used planners, CEM and model predictive path integral (MPPI) control~\cite{williams2017mppi}.
The reason is that the planner keeps the candidates with the lowest predicted cost.
A candidate can have a low predicted cost because it is good, or because the model underestimates its cost~\cite{smith2006optimizer,you2026control}.
The longer the planner searches, the more candidates of the second kind it can keep.

Earlier work observed that an optimizer can exploit the errors of a learned model~\cite{henaff2017model,ha2018recurrent,boney2019regularizing,janner2021trajectory,janner2022diffuser}.
On latent world models, CEM has been described as comparatively robust to model error~\cite{agrawal2025closing,terver2025jepawms}.
Most benchmarks report the success of an episode with replanning, which does not show how often a single search ends in an adversarial plan.
Novelty scores for latent world models judge a plan by its predicted trajectory~\cite{wmvae2025}, which the planner has optimized to look as if it reaches the goal, whether the plan is adversarial or genuine.
To check that predicted trajectories are realizable, ACID~\cite{acid2026} adds a clever planning cost, in which an inverse-dynamics network infers the actions from the predicted states and the planner penalizes planned actions that differ from them.
The world models we study predict one state for every block of five actions, and many blocks connect the same two states.

\begin{figure*}[t]
  \centering
  \includegraphics[width=\linewidth]{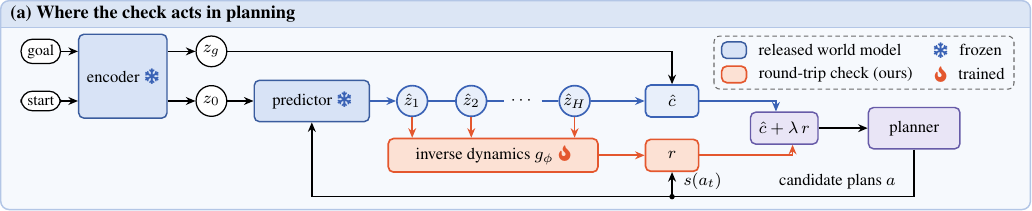}\par\vspace{4pt}
  \includegraphics[width=0.7216\linewidth]{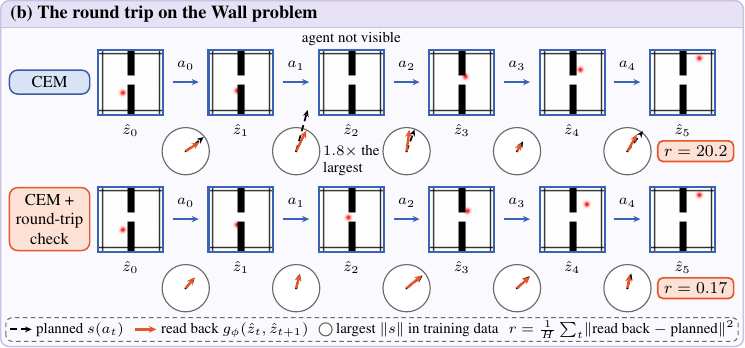}\hfill\includegraphics[width=0.2611\linewidth]{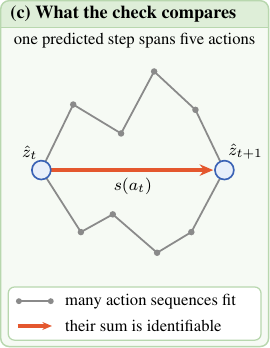}%
  {\phantomsubcaption\label{fig:method:pipe}}%
  {\phantomsubcaption\label{fig:method:a}}%
  {\phantomsubcaption\label{fig:method:b}}%
  \caption{\textbf{The round-trip check.}
  (a) The frozen predictor rolls out the candidate plans $a$ of the planner, and the inverse-dynamics network $g_\phi$ reads the summed action of each block back from consecutive predicted states.
  Its residual $r$ against the planned sums $s(a_t)$ is added to the predicted cost $\hat c$ with weight $\lambda$.
  (b) Decoded predicted states of the final plan of CEM and of the plan found with the check, with the residual $r$ of each plan at the end of its row.
  Each circle compares the planned sum of a block with the sum that $g_\phi$ reads back, and its radius is the largest sum in the training data.
  The final plan of CEM asks for block sums that its own predicted states do not support, while the plan found with the check agrees with its prediction.
  (c) Many action sequences connect the same two predicted states, and only their sum is identifiable, so the check reads back this sum.}
  \label{fig:method}
\end{figure*}

We measure these adversarial plans and show that scores computed from the model's prediction do not separate them from genuine ones (\cref{sec:phenomenon,sec:selfcert}).
For a planner as a whole, we propose a test that runs the planner once and executes its plan after every iteration (\cref{sec:test}).
Among the problems that some iteration solves, the fraction that the final plan fails is the \emph{adversarial rate} of the planner, and a planner with a high adversarial rate needs repair.
Unlike an adversarial image, a plan can be checked against its own prediction, so the world model needs no retraining.
The two states at the ends of a block determine only the sum of its actions.
On TwoRoom, a network that sees only the two states predicts this sum almost perfectly, but predicts the single actions of the block no better than their mean (\cref{sec:closer}).
We therefore repair the planner with a small inverse-dynamics network that reads this sum back from the predicted states, and penalize plans whose actions do not add up to it.
As \Cref{fig:method} shows, the actions go to predicted states and back, so we call the method the \emph{round-trip check} (\cref{sec:certificate}).
Its network trains in minutes on one GPU and adds only \ovhWall\% to \ovhCube\% to the planning time.

With the check, the success of CEM on Wall rises from \wallBase{} to \wallOurs{} after all 30 iterations, without early stopping and without selecting among plans afterwards (\cref{sec:experiments}).
CEM alone stays below it even when stopped at its best iteration (\cref{sec:closer}).
The check also improves CEM on Cube with LeWM, without and with replanning, and it can be applied to other world models and planners (\cref{sec:plugin}).
On JEPA-WM~\cite{terver2025jepawms}, it raises the success of MPPI from 0.56 to 0.94 and lifts it above CEM.
On Wall, the adversarial plans exploit position labels in the released data that disagree with the frames (\cref{sec:wall}).
Without correcting these labels, the check repairs the planner and reaches \wallOurs{}, close to the \wallFixedAll{} of a model fine-tuned on corrected labels.

\noindent In summary, our contributions are:
\begin{itemize}
  \item We find that a longer search by CEM or MPPI on released latent world models can end in adversarial plans, on up to half of the problems that CEM had solved, and that the model cannot recognize them. To our knowledge, we are the first to measure this effect systematically.
  \item We propose a simple test that tells whether a planner needs repair. For the repair, our round-trip check reads the sum of each block of actions back from the predicted states and penalizes plans whose actions do not add up to it.
  \item On three released world models, the check improves CEM and MPPI without retraining and at only \ovhWall\% to \ovhCube\% more planning time. On Wall, it repairs the planner even though the training labels disagree with the frames.
\end{itemize}

\section{Related work}
\label{sec:related}

\parahead{Planning with latent world models.}
DINO-WM~\cite{zhou2025dinowm} plans with CEM in the patch-feature space of a frozen DINOv2~\cite{oquab2024dinov2} encoder, and
PLDM~\cite{sobal2025pldm}, V-JEPA~2-AC~\cite{assran2025vjepa2}, LeWM~\cite{lewm2026} and JEPA-WMs~\cite{terver2025jepawms} plan in the same way with other encoders, predictors, objectives and planners.
Recent work asks how many steps ahead such models can plan~\cite{planninglimits2026} and makes the latent cost a better guide for planning~\cite{scale2026,djepa2026}.
Most of them evaluate a planner by the success of an episode with replanning.
We execute the plan after every iteration of a single search, which tells whether a planner needs repair.

\parahead{Adversarial examples and model exploitation.}
An input optimized against a classifier can be assigned to a wrong class with high confidence~\cite{szegedy2014intriguing,goodfellow2015explaining,nguyen2015fooled}.
Later work makes these attacks stronger~\cite{moosavi2016deepfool,carlini2017towards,madry2018towards} and carries them into the physical world~\cite{kurakin2017physical,eykholt2018robust}.
Detectors and defenses can be bypassed by attacks designed against them~\cite{carlini2017detected,athalye2018obfuscated}.
Optimizing plans or policies against a learned dynamics model has been described as such an attack~\cite{henaff2017model,ha2018recurrent,boney2019regularizing,janner2021trajectory,janner2022diffuser}.
Optimizing a policy against a learned reward model~\cite{gao2023scaling} or a design against a learned objective~\cite{trabucco2021coms} likewise overoptimizes a proxy.
With a state-based model, Boney \etal~\cite{boney2019regularizing} show a case in which the true return falls over planning iterations while the predicted return stays high.
On latent world models, LEAP~\cite{leap2026} notes that planners can lower the latent cost while leaving the data distribution, and CEM has been called comparatively robust to model error~\cite{agrawal2025closing,terver2025jepawms}.
We measure how often CEM and MPPI end in adversarial plans on released latent world models, which cannot recognize them.

\begin{figure*}[t]
  \centering
    \includegraphics[width=0.858\linewidth]{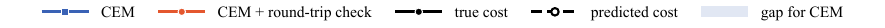}\\[2pt]
    \begin{subfigure}[t]{0.245\linewidth}
      \centering
      \includegraphics[width=\linewidth]{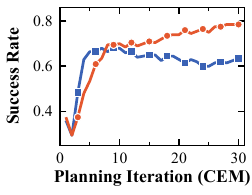}
      \caption{Success rate on Cube}
      \label{fig:curves:a}
    \end{subfigure}\hfill
    \begin{subfigure}[t]{0.245\linewidth}
      \centering
      \includegraphics[width=\linewidth]{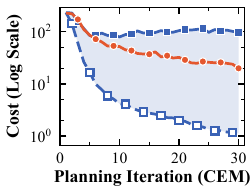}
      \caption{Cost on Cube}
      \label{fig:curves:b}
    \end{subfigure}\hfill
    \begin{subfigure}[t]{0.245\linewidth}
      \centering
      \includegraphics[width=\linewidth]{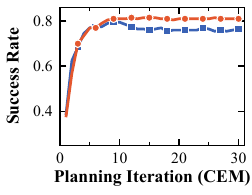}
      \caption{Success rate on TwoRoom}
      \label{fig:curves:c}
    \end{subfigure}\hfill
    \begin{subfigure}[t]{0.245\linewidth}
      \centering
      \includegraphics[width=\linewidth]{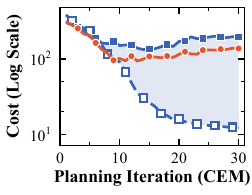}
      \caption{Cost on TwoRoom}
      \label{fig:curves:d}
    \end{subfigure}%
  \caption{\textbf{The same divergence for CEM with LeWM.}
  We use 200 problems per task.
  (a, c) Success rate of the plan executed after each iteration.
  (b, d) Median cost of that plan, dashed as the model predicts it and solid as the simulator measures it.
  On both tasks the predicted cost of the plan of CEM keeps falling, while its true cost stops falling or rises, so the shaded gap between the two widens.
  With the round-trip check, the plan ends at a lower true cost and a higher success rate.}
  \label{fig:curves}
\end{figure*}
\parahead{Remedies and consistency checks.}
Model-based reinforcement learning plans with model ensembles~\cite{chua2018pets} and penalizes uncertain regions~\cite{yu2020mopo,kidambi2020morel}.
Boney \etal~\cite{boney2019regularizing} regularize plans with a denoising autoencoder over state-action trajectories, and Agrawal \etal~\cite{agrawal2025closing} fine-tune a latent world model on synthesized and adversarially perturbed trajectories.
Ruan \etal~\cite{ruan2026future} diagnose the rollouts of world action models by action-state consistency.
Inverse-dynamics models have long been used in motor learning and robotics~\cite{jordan1992forward,agrawal2016poke,christiano2016transfer}, in curiosity-driven exploration~\cite{pathak2017curiosity}, in imitation from observation~\cite{torabi2018bco,baker2022vpt} and to execute generated videos~\cite{du2023unipi}.
Closest to our repair, ACID~\cite{acid2026} adds the residual of such a model to the planning cost, a clever design with reported gains across four world models.
We first analyze the failed plans as adversarial examples of the world model and derive a test of whether a planner needs repair.
ACID assumes that two consecutive states determine the action between them, and with one state per action our check has the same target.
With one state per block of five actions, the two states determine only the sum of the block, which our check reads back.

\begin{table}[t]
  \centering
  \caption{\textbf{Clipping the actions removes only part of the adversarial plans.}
  Success rate of the final plan of CEM after 30 iterations.
  The adversarial rate is the fraction of the problems solved by the plan of some iteration that the final plan fails.
  Clipping limits every sampled action to the range of the training data.}
  \label{tab:phenomenon}
  \small
  \setlength{\tabcolsep}{4.5pt}
  \begin{tabular}{@{}llccc@{}}
    \toprule
    Measure & Plan of CEM & Wall & Cube & TwoRoom \\
    \midrule
    \multirow{2}{*}{Success} & final                   & \wallBase{} & \cubeBase{} & \twoBase{} \\
                             & + action clipping       & 0.52 & 0.61 & 0.70 \\
    \midrule
    \multirow{2}{*}{\shortstack[l]{Adversarial\\rate}} & final & 52\% & 16\% & 7\% \\
                             & + action clipping       & 37\% & 16\% & 1\% \\
    \bottomrule
  \end{tabular}
\end{table}
\section{Planning as an adversarial search}
\label{sec:analysis}

We find that a longer search by a planner on a latent world model can end in adversarial plans (\cref{sec:phenomenon}), and that the model cannot recognize them (\cref{sec:selfcert}).
Our test finds this for CEM and MPPI on three world models (\cref{sec:test}).

\subsection{Setup}
\label{sec:setup}
We study the released checkpoints of DINO-WM~\cite{zhou2025dinowm} on Wall and of LeWM~\cite{lewm2026} on Cube~\cite{park2025ogbench} and TwoRoom~\cite{sobal2025pldm}.
Wall and TwoRoom are navigation through a door between two rooms, and Cube is manipulation of a cube by a robot arm.
Each problem is a start and a goal drawn by the released evaluation, 148 on Wall and 200 on the other tasks.
Both models predict one latent state per \emph{block} of 5 environment actions, and a plan spans 5 blocks.
The planner is the released CEM, which samples 300 action sequences in each of 30 iterations and refits its distribution to the 30 with the lowest cost.
The released planners return the mean of this distribution as their plan~\cite{zhou2025dinowm,lewm2026}, and we execute it after every iteration from the initial state.
The \emph{predicted cost} $\hat c$ is the planner's own objective for this plan, and the \emph{true cost} $c$ is this objective for the frame that the simulator reaches.
\subsection{More planning, worse plans}
\label{sec:phenomenon}

\parahead{A longer search turns genuine plans into adversarial ones.}
One would expect a longer search to find a better plan.
It does lower the predicted cost, by a factor of 30 to 200 on all three tasks.
Yet the true cost reaches a minimum early and then rises, as \Cref{fig:teaser} shows for Wall and \Cref{fig:curves} for Cube and TwoRoom.
On Wall the success peaks at \wallBasePeak{} after \wallBasePeakIt{} iterations and falls to \wallBase{} after 30 and to 0.28 after 60.
On Cube it declines from 0.68 to \cubeBase{}, and on TwoRoom from \twoBasePeak{} to \twoBase{}.
The final plan thus fails 52\% of the problems that an earlier plan had solved on Wall, 16\% on Cube and 7\% on TwoRoom.
Since the model predicts a lower cost for these failing plans, the search has turned genuine plans into adversarial ones.

\parahead{Keeping the actions in range is not enough.}
Some adversarial plans use actions outside the range of the training data, and an obvious remedy is to clip every sampled action to this range.
As \Cref{tab:phenomenon} shows, clipping lowers the adversarial rate from 52\% to 37\% on Wall and from 7\% to 1\% on TwoRoom, but it lowers success on Cube and TwoRoom.
On Cube, where only 3\% of the actions leave the range, the adversarial rate stays at 16\%.
These adversarial plans keep their actions within the range and combine them into a sequence that the model mispredicts.
No limit on the actions can exclude such plans, so a repair must catch them as well.
\subsection{The model cannot recognize adversarial plans}
\label{sec:selfcert}

A natural remedy is to detect adversarial plans.
Novelty scores for latent world models~\cite{wmvae2025} judge a plan by its predicted trajectory.
But the planner has optimized exactly this trajectory to look as if it reaches the goal, whether the plan is adversarial or genuine.
A score computed from the prediction should thus see little difference between them.

\definecolor{tblue}{HTML}{3A62B5}
\definecolor{torange}{HTML}{E4572E}
\newcommand{\hp}[2]{{\setlength{\fboxsep}{1.8pt}\colorbox{tblue!#1}{\makebox[2.2em][c]{#2}}}}
\newcommand{\ha}[2]{{\setlength{\fboxsep}{1.8pt}\colorbox{torange!#1}{\makebox[2.2em][c]{#2}}}}
\begin{table}[t]
  \centering
  \caption{\textbf{No score reliably tells a plan that fails from one that succeeds.}
  AUROC on the problems that an earlier iteration of CEM had solved.
  Higher is better, with 1 for a perfect separation and 0.5 for a random guess.
  A darker cell is further above 0.5.}
  \label{tab:detectors}
  \small
  \setlength{\tabcolsep}{6pt}
  \begin{tabular}{@{}llcc@{}}
    \toprule
    Input & Score & Wall & Cube \\
    \midrule
    \multirow{3}{*}{Prediction} & predicted cost $\hat c$   & \hp{2}{0.51}  & \hp{9}{0.55}  \\
                                & nearest-neighbor distance~\cite{sun2022knn} & \hp{0}{0.47}  & \hp{0}{0.46}  \\
                                & best other OOD score  & \hp{13}{0.56} & \hp{27}{0.63} \\
    \midrule
    \multirow{2}{*}{Actions}    & largest action            & \ha{23}{0.61} & \ha{1}{0.50}  \\
                                & share of out-of-range actions & \ha{26}{0.63} & \ha{12}{0.56} \\
    \bottomrule
  \end{tabular}
\end{table}

On the problems that CEM solves at some iteration, the plan of every later iteration either still succeeds or has turned adversarial.
\Cref{tab:detectors} gives the area under the receiver operating characteristic curve (AUROC) for telling the two apart, where 0.5 is a random guess.
The upper three scores use only the prediction.
The predicted cost minimized by the planner is about as low for adversarial plans as for genuine ones.
Sun \etal~\cite{sun2022knn} detect out-of-distribution (OOD) inputs by the distance to their nearest neighbor in the training data.
This score stays near 0.5 when applied to the last predicted state, because the predicted end of an adversarial plan lies as close to the training data as a real frame does, for all but one of them on Wall and for 95\% on Cube.
The model thus predicts a realistic future for a plan that fails.
The third row is the best of further OOD scores~\cite{wang2022vim,lee2018simple,wmvae2025} for each model, and it reaches at most 0.63.

We add two scores computed from the planned actions.
They are the largest absolute action and the share of actions outside the range of the training data.
On Wall, where adversarial plans contain out-of-range actions, they reach 0.61 and 0.63, better than the prediction alone but far from a reliable separation.
On Cube, where adversarial plans keep their actions within the range, they reach at most 0.56.
Like detectors of adversarial images~\cite{carlini2017detected}, no score therefore separates adversarial from genuine plans reliably.
\subsection{A test of whether a planner needs repair}
\label{sec:test}
No score tells whether a single plan is adversarial, but executing the plans of a planner shows how often its search ends in one.
We run the planner once on each problem with its usual budget and execute its plan after every iteration.
If the plan of some iteration solves a problem, the search has found a genuine plan.
If the final plan then fails, the search has given it up for an adversarial plan that the model rates better.
On every problem where the two can be compared, 61 on Wall and 21 on Cube, the failing plan has the lower predicted cost.
The \emph{adversarial rate} of a planner is the fraction of the problems solved at some iteration that its final plan fails.
The test needs no training or change to the planner, and it can be run on about 50 problems of a separate seed first.

\begin{table}[t]
  \centering
  \caption{\textbf{The seven planners above the line need repair.}
  Problems that the planner solves at some iteration and those among them that its final plan fails.
  The adversarial rate is the ratio of the two.
  For MPPI the test executes the mean plan.}
  \label{tab:test}
  \small
  \setlength{\tabcolsep}{4.0pt}
  \begin{tabular}{@{}lllccc@{}}
    \toprule
    \multirow{2}{*}{\hstackl{World}{model}} & \multirow{2}{*}{Task} & \multirow{2}{*}{Planner} & \multicolumn{2}{c}{Problems} & \multirow{2}{*}{\hstackc{Adversarial}{rate}} \\
    \cmidrule(lr){4-5}
    & & & Solved & Failed & \\
    \midrule
    DINO-WM & Wall      & MPPI & 111 & 75 & 68\% \\
    DINO-WM & Wall      & CEM  & 122 & 64 & 52\% \\
    LeWM    & Reacher   & CEM  & 137 & 64 & 47\% \\
    JEPA-WM & Wall      & MPPI & 110 & 35 & 32\% \\
    LeWM    & Cube      & MPPI & 151 & 25 & 17\% \\
    LeWM    & Cube      & CEM  & 152 & 25 & 16\% \\
    JEPA-WM & Wall      & CEM  & 104 & 15 & 14\% \\
    \midrule
    LeWM    & PushT     & CEM  & 178 & 15 & \phantom{0}8\%  \\
    LeWM    & TwoRoom   & CEM  & 165 & 12 & \phantom{0}7\%  \\
    LeWM    & TwoRoom   & MPPI & 185 & 13 & \phantom{0}7\%  \\
    JEPA-WM & PushT     & CEM  & \phantom{0}48  & \phantom{0}3  & \phantom{0}6\%  \\
    DINO-WM & PushT     & CEM  & \phantom{0}46  & \phantom{0}0  & \phantom{0}0\%  \\
    DINO-WM & PointMaze & CEM  & \phantom{0}50  & \phantom{0}0  & \phantom{0}0\%  \\
    \bottomrule
  \end{tabular}
\end{table}

\Cref{tab:test} applies the test to thirteen combinations of world model, task and planner.
They cover every task of the DINO-WM and LeWM releases that is scored by success, which adds PushT~\cite{chi2023diffusion}, PointMaze~\cite{fu2020d4rl} and Reacher~\cite{tassa2018dmcontrol} to the tasks of \cref{sec:setup}, and a third released world model, JEPA-WM~\cite{terver2025jepawms}.
The combinations fall into two groups.
Seven of them lose 14\% to 68\% of the problems they solve.
They include both planners and all three world models, so adversarial plans are not specific to one planner or one model, and these planners need repair.
The other six lose at most 8\% and mostly end in genuine plans, so their planners are sound.
The test thus finds that more than half of the planners we test need repair.
\section{The round-trip check}
\label{sec:certificate}

\parahead{Reading the actions back.}
\Cref{sec:analysis} shows that a longer search can turn genuine plans into adversarial ones, and that the model cannot recognize them.
A repair need not recognize them either.
In each iteration the planner only ranks the candidates of one problem, so a penalty that tends to be larger for adversarial candidates steers it away from them.
The predicted cost cannot be this penalty, since the planner already minimizes it.
Unlike an image, a plan can be checked against its own prediction, because every planned action must match the transition that the model predicts for it.
We therefore read the actions back with an inverse-dynamics model (IDM)~\cite{jordan1992forward,agrawal2016poke,christiano2016transfer} $g_\phi(z, z')$, which predicts the action moving the latent state from $z$ to $z'$.
We apply it to each predicted transition $(\hat z_t, \hat z_{t+1})$ of a candidate plan and compare the action it reads back with the planned one.
We call the path from planned actions to predicted states and back the \emph{round trip}, which \Cref{fig:method:a} shows.

\parahead{The target must be identifiable.}
The world models that we study predict one latent step per block $a_t$ of $k{=}5$ environment actions.
Many blocks lead from one latent state to the same next state, so the two states do not determine the $k$ actions between them, as \Cref{fig:method:b} illustrates.
The squared residual of any per-action IDM $\tilde g$ splits exactly into two parts, $\lVert \tilde g - a_t\rVert^2 = \tfrac1k \lVert s(\tilde g) - s(a_t)\rVert^2 + \lVert \tilde g^{\perp} - a_t^{\perp}\rVert^2$, where $s$ sums the actions of a block and ${}^{\perp}$ subtracts their mean from each.
If the next state depends on a block only through its sum and determines it, the first part can be zero on all the data, while the second penalizes genuine plans by at least $\mathbb E\bigl[\mathrm{tr}\,\mathrm{Cov}(a_t \mid z_t, s(a_t))\bigr]$ in expectation, the spread of the actions among blocks with the same sum from the same state, whatever the network.
We keep the first part and compare the summed action of each block.

\parahead{The round-trip check.}
Where the two states determine the sum, least squares recovers it and a small network suffices.
The IDM is a multilayer perceptron (MLP) with two hidden layers of 512 units and GELU activations~\cite{hendrycks2016gelu} on $[z_t,\, z_{t+1}-z_t]$, trained on consecutive encoded frames of the world model's training data $\mathcal D$, where $a_t$ is the block between them.
The \emph{round-trip check} applies it to the predicted states of a candidate and adds the residual $r(a)$ to the planning cost, as \Cref{fig:method:pipe} shows,
\begin{equation}
\begin{aligned}
  \mathcal L(\phi) &= \mathbb E_{(z_t, a_t, z_{t+1}) \sim \mathcal D}\, \bigl\lVert g_\phi(z_t, z_{t+1}) - s(a_t) \bigr\rVert^2, \\
  r(a) &= \frac1H \sum_{t=0}^{H-1} \bigl\lVert g_\phi(\hat z_t, \hat z_{t+1}) - s(a_t) \bigr\rVert^2, \\
  \tilde c(a) &= \hat c(a) + \beta\, \frac{\sigma(\hat c)}{\sigma(r)}\, r(a),
\end{aligned}
\label{eq:cert}
\end{equation}
where $H$ is the number of blocks, $a_t$ is the $t$-th block of the plan and $\sigma(\cdot)$ is the standard deviation over the candidates of the iteration.
We normalize by this ratio as ACID~\cite{acid2026} does, so that the penalty's effect on the ranking is independent of scale, and $\beta$ is the only hyperparameter.

\parahead{Why and when the check works.}
Let $a^{\mathrm{gen}}$ be the best genuine plan and $a^{\mathrm{adv}}$ the best adversarial plan, and suppose that the model prefers the latter, so that $\hat c(a^{\mathrm{adv}}) < \hat c(a^{\mathrm{gen}})$.
The penalized planner prefers the genuine plan when
\begin{equation}
  \lambda\,\bigl[r(a^{\mathrm{adv}}) - r(a^{\mathrm{gen}})\bigr] \;>\; \hat c(a^{\mathrm{gen}}) - \hat c(a^{\mathrm{adv}}),
  \label{eq:helps}
\end{equation}
where $\lambda = \beta\,\sigma(\hat c)/\sigma(r)$ is the weight in \cref{eq:cert}.
In words, the adversarial plan must be less consistent with its own prediction than the genuine plan, by more than the model's mistaken preference.
The residual $r$ therefore only has to move the selection away from adversarial plans.
The check is a minimal repair for the planners that need it (\cref{sec:test}), and its target fits tasks whose actions add up within a block, as on Wall, TwoRoom and Cube.
On Reacher, whose joint torques act through richer dynamics, the check still raises success from 0.37 to 0.41 (\cref{sec:plugin}).
\section{Experiments}
\label{sec:experiments}

\begin{table}[t]
  \centering
  \caption{\textbf{Success rate of the executed plan} with CEM on DINO-WM (Wall) and LeWM (Cube and TwoRoom).
  Without replanning, the final plan after 30 iterations is executed.
  With replanning, the agent follows the released model predictive control (MPC) protocol of each benchmark.}
  \label{tab:main}
  \small
  \setlength{\tabcolsep}{3pt}
  \adjustbox{max width=\linewidth}{%
  \begin{tabular}{lcccccc}
    \toprule
    & \multicolumn{3}{c}{Without replanning} & \multicolumn{3}{c}{With replanning (MPC)} \\
    \cmidrule(lr){2-4}\cmidrule(l){5-7}
    Method & Wall & Cube & TwoRoom & Wall & Cube & TwoRoom \\
    \midrule
    CEM                               & 0.39 & 0.64 & 0.77 & \relWallBase{} & 0.70 & 0.85 \\
    + per-action                      & 0.64 & 0.71 & 0.78 & \relWallAcid{} & 0.75 & 0.83 \\
    \rowcolor{black!8}
    + round-trip (ours)               & \textbf{0.76} & \textbf{0.79} & \textbf{0.81} & \textbf{\relWallOurs} & \textbf{0.81} & \textbf{0.87} \\
    \bottomrule
  \end{tabular}}
\end{table}

\definecolor{qblue}{HTML}{3A62B5}
\definecolor{qorange}{HTML}{E4572E}
\definecolor{qgreen}{HTML}{3EA44E}
\newcommand{\qfrc}[3]{{\setlength{\fboxsep}{0pt}\setlength{\fboxrule}{#2}\fcolorbox{#1}{white}{#3}}}
\newcommand{\qmk}[2]{\llap{\raisebox{2.2pt}{{\setlength{\fboxsep}{0.6pt}\colorbox{white}{\textcolor{#1}{\footnotesize\ding{#2}}}}}\kern2.6pt}}
\newcommand{\qG}[1]{\qfrc{black!45}{0.4pt}{#1}}
\newcommand{\qP}[1]{\qfrc{qblue}{1.1pt}{#1}\qmk{qblue}{51}}
\newcommand{\qE}[1]{\qfrc{black}{1.1pt}{#1}\qmk{black}{55}}
\newcommand{\qK}[1]{\qfrc{qorange}{1.1pt}{#1}\qmk{qorange}{51}}
\newcommand{\qleg}[2]{\textcolor{#1}{\rule[0.45ex]{1.5em}{1.4pt}}\,\,#2}
\begin{figure*}[t]
  \centering
  \footnotesize
  \setlength{\tabcolsep}{1.5pt}
  \begin{subfigure}[b]{0.56\linewidth}
    \centering
    \begin{tabular}{@{}ccccc@{}}
      Goal & Predicted & Executed & With the check & Paths \\[1pt]
      \qG{\includegraphics[width=0.176\linewidth]{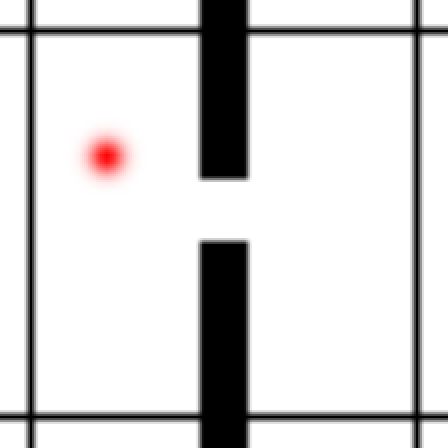}} & \qP{\includegraphics[width=0.176\linewidth]{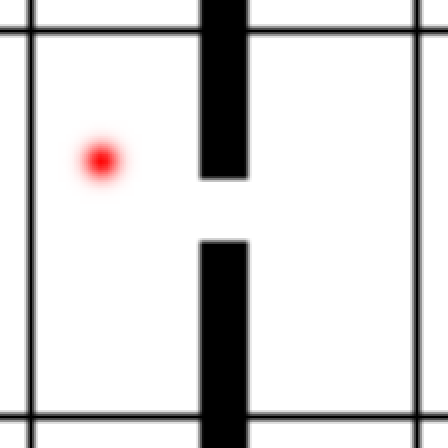}} & \qE{\includegraphics[width=0.176\linewidth]{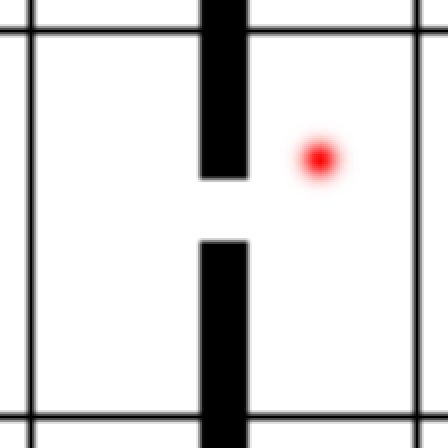}} & \qK{\includegraphics[width=0.176\linewidth]{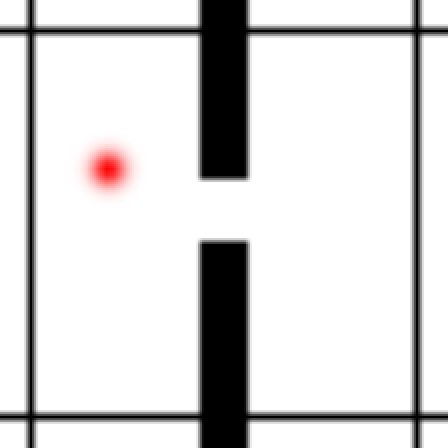}} & \qG{\includegraphics[width=0.176\linewidth]{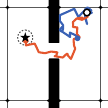}} \\
    \end{tabular}
    \caption{Wall with JEPA-WM and MPPI}
    \label{fig:qual:g}
  \end{subfigure}\hfill
  \begin{subfigure}[b]{0.43\linewidth}
    \centering
    \begin{tabular}{@{}cccc@{}}
      Goal & Predicted & Executed & With the check \\[1pt]
      \qG{\includegraphics[width=0.221\linewidth]{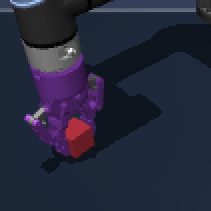}} & \qP{\includegraphics[width=0.221\linewidth]{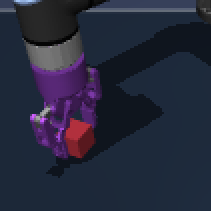}} & \qE{\includegraphics[width=0.221\linewidth]{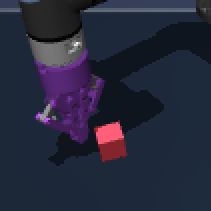}} & \qK{\includegraphics[width=0.221\linewidth]{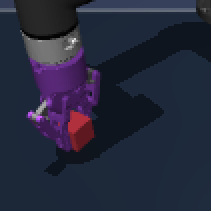}} \\
    \end{tabular}
    \caption{Cube with LeWM and CEM}
    \label{fig:qual:b}
  \end{subfigure}\\[4pt]
  \qleg{qblue}{path of the planner alone}\qquad
  \qleg{qorange}{path with the round-trip check}\qquad
  $\boldsymbol{\circ}$\,\,start\qquad $\bigstar$\,\,goal
  \caption{\textbf{The standard planners can end in adversarial plans, and the round-trip check finds plans that reach the goal.}
  Each example shows the goal, the end of the final plan as the model predicts it and as it is executed, and the end of the plan found with the round-trip check.
  The predicted end is the nearest training frame, and a tick or a cross marks whether an end reaches the goal.
  In (a) the final plan of MPPI fails, and in (b) CEM solves the problem after 10 iterations and loses it after 30.
  The dotted circle marks the distance within which a problem counts as solved.}
  \label{fig:qual}
\end{figure*}

\subsection{Experimental setup}
\label{sec:expsetup}
For a fair comparison, we measure all numbers ourselves on the same problems.
To compare two methods, we count the problems that only one of them solves and apply a two-sided sign test to them.
Its $p$-value is the probability of an imbalance at least this large if both methods were equally good.
With replanning, the agent follows the released MPC protocol of each benchmark.
The Wall protocol does not limit the number of plans, so we report its success after the second plan, as the LeWM protocol executes two.
The per-action baseline uses the target of ACID~\cite{acid2026} at its reported weights of 0.07 on LeWM and 0.5 on DINO-WM, with the same network and data as the check.
The weight $\beta$ is chosen among 0.25, 0.5, 1 and 2 on about 50 separate problems per task and protocol.
The IDM has at most \idmShareCube\% of the predictor's parameters, trains in \idmTrainWall{} minutes on Wall and \idmTrainCube{} on Cube, and adds \ovhWall\% and \ovhCube\% to planning time.

\subsection{Repairing adversarial plans}
\label{sec:results}
\Cref{tab:test} shows that CEM needs repair on Wall and Cube.

\parahead{Without replanning.}
As \Cref{tab:main} shows, the check nearly doubles the success of the final plan on Wall, from \wallBase{} to \wallOurs{}, and raises it on Cube from \cubeBase{} to \cubeOurs{} (both $p<10^{-4}$).
On Wall it lowers the adversarial rate of CEM from 52\% to 10\%.
\Cref{fig:teaser,fig:curves} show that the check lowers the true cost on every task, and \Cref{fig:qual:b} shows a repaired plan.
Clipping cannot reach adversarial plans within the range. It gives 0.52, 0.61 and 0.70, below the check on all three tasks and below CEM alone on Cube and TwoRoom.

\parahead{With replanning.}
MPC lets the agent recover from a bad plan, and the check acts inside each search.
On Cube it raises success from \clCubeBase{} to \clCubeOurs{} ($p=3\times10^{-4}$).
The released Wall protocol stops each search after 10 iterations, before the plans of CEM degrade (\Cref{fig:teaser:a}).
CEM then recovers with its second plan to \relWallBase{}, against \relWallOurs{} with the check.
The check needs fewer plans: its first plan already solves 0.66 of the problems, against 0.54 for CEM alone ($p=0.01$).

\subsection{Other planners in need of repair}
\label{sec:plugin}
\begin{table}[t]
  \centering
  \caption{\textbf{The other planners in need of repair.}
  Success rate of the final plan without replanning, for the planner alone and with the check.
  The adversarial rate is that of the planner alone.
  JEPA-WM uses 126 Wall problems. MPPI executes a sampled plan, as in its release. The mean plan of \Cref{tab:test} solves 0.60 on JEPA-WM.}
  \label{tab:plugin}
  \small
  \setlength{\tabcolsep}{5pt}
  \begin{tabular}{@{}lllccc@{}}
    \toprule
    \multirow{2}{*}{\hstackl{World}{model}} & \multirow{2}{*}{Task} & \multirow{2}{*}{Planner} & \multirow{2}{*}{\hstackc{Adversarial}{rate}} & \multicolumn{2}{c}{Success} \\
    \cmidrule(l){5-6}
    & & & & Alone & Check \\
    \midrule
    \multirow{2}{*}{JEPA-WM} & \multirow{2}{*}{Wall} & CEM  & 14\% & 0.71 & \textbf{0.75} \\
                             &                       & MPPI & 32\% & 0.56 & \textbf{0.94} \\
    \midrule
    DINO-WM & Wall & MPPI & 68\% & 0.24 & \textbf{0.55} \\
    \midrule
    \multirow{2}{*}{LeWM} & Cube    & MPPI & 17\% & 0.62 & \textbf{0.72} \\
                          & Reacher & CEM  & 47\% & 0.37 & \textbf{0.41} \\
    \bottomrule
  \end{tabular}
\end{table}

\begin{table}[t]
  \centering
  \caption{\textbf{Planners with a low adversarial rate.}
  Success rate of the final plan without replanning, for the planner alone and with the round-trip check.
  MPPI executes a sampled plan. $^\dagger$\,With re-estimated BatchNorm statistics.}
  \label{tab:noneed}
  \small
  \setlength{\tabcolsep}{3.8pt}
  \begin{tabular}{@{}lllccc@{}}
    \toprule
    \multirow{2}{*}{\hstackl{World}{model}} & \multirow{2}{*}{Task} & \multirow{2}{*}{Planner} & \multirow{2}{*}{\hstackc{Adversarial}{rate}} & \multicolumn{2}{c}{Success} \\
    \cmidrule(l){5-6}
    & & & & Alone & Check \\
    \midrule
    \multirow{4}{*}{LeWM} & PushT                    & CEM  & \phantom{0}8\%  & \textbf{0.82} & 0.81 \\
                          & \multirow{2}{*}{TwoRoom} & CEM  & \phantom{0}7\%  & \twoBase{} & \textbf{0.81} \\
                          &                          & MPPI & \phantom{0}7\%  & \textbf{0.86} & 0.80 \\
                          & TwoRoom$^\dagger$        & CEM  & 10\% & \bnCemOpen{} & \textbf{\bnCheckOpen{}} \\
    \midrule
    JEPA-WM & PushT & CEM & \phantom{0}6\% & \textbf{0.67} & 0.52 \\
    \midrule
    \multirow{2}{*}{DINO-WM} & PushT     & CEM & \phantom{0}0\% & \textbf{0.92} & 0.88 \\
                             & PointMaze & CEM & \phantom{0}0\% & \textbf{1.00} & 0.92 \\
    \bottomrule
  \end{tabular}
\end{table}

\Cref{tab:plugin} covers the other five planners that need repair according to \Cref{tab:test}.
The check uses only the latent states that the planner already predicts, so it carries over to them unchanged.
On JEPA-WM~\cite{terver2025jepawms}, whose released evaluation executes one plan per episode, we train the IDM on its latents.
We run its released CEM configuration and its MPPI planner on Wall at the same budget, with $\beta=1$.
As \Cref{tab:plugin} shows, the check raises the success of MPPI from 0.56 to 0.94 and that of CEM from 0.71 to 0.75.
\Cref{fig:qual:g} shows a problem that MPPI alone fails.
The model predicts that its final plan reaches the goal, but the executed path never passes the door, while the plan found with the check goes through it to the goal.
With MPPI ported to DINO-WM and LeWM, the check also raises success on Wall and Cube, and on all three models MPPI with the check beats CEM alone.

\subsection{Where the search ends in genuine plans}
\label{sec:noneed}

The six planners in the lower part of \Cref{tab:test} lose at most 8\% of the problems they solve.
\Cref{tab:noneed} shows that the check brings no significant gain on any of them, so they are best left as they are.
Re-estimating the BatchNorm~\cite{ioffe2015batchnorm} statistics of the released TwoRoom checkpoint lowers its validation loss from \bnLossReleased{} to \bnLossRecal{}.
It also raises the adversarial rate of CEM from 7\% to 10\%, and the gain of the check grows with it, from 0.04 to 0.09 ($p<10^{-3}$).
Over all fourteen settings, the gain of the check grows with the adversarial rate (Spearman correlation 0.85), so the test tells where to use the check, which raises the success of all seven planners in need of repair, five of them significantly.
For the ten combinations with separate problems, the test on these problems gives the same two groups, with rates above 10\% and below 8\%.

\subsection{Analysis}
\label{sec:closer}\label{sec:wall}

\parahead{The check repairs the planner without corrected labels.}
The adversarial rate is highest on Wall, so we trace it to its cause.
In the released Wall data, the agent moves two units per unit of action in the frames but only one in the stored positions.
The model learns both rates with a normal validation loss and underestimates by half how far an action moves the position.
The adversarial plans exploit this error with actions that are too large.
As \Cref{tab:walllabels} shows, fine-tuning the model on corrected positions lowers the adversarial rate of CEM from 52\% to 5\% and raises its success from \wallBase{} to \wallFixedAll{}.
The same fine-tuning on the original labels reaches only 0.42 on 50 problems, against \wallFixed{} with corrected labels.
The round-trip check needs no such correction.
Without retraining or knowledge of the labels, it lowers the adversarial rate to 10\% and reaches \wallOurs{}.
This closes 84\% of the gap to the corrected model, and the rest is not significant (${p=0.12}$).
\begin{table}[t]
  \centering
  \caption{\textbf{Wall with corrected labels.}
  Success rate of the final plan of CEM on the 148 problems and its adversarial rate.
  The last two models fine-tune the released one on position labels corrected with the simulator or read from the frames, and plan without the check.}
  \label{tab:walllabels}
  \small
  \setlength{\tabcolsep}{6pt}
  \begin{tabular}{lcc}
    \toprule
    Model & Success & Adversarial rate \\
    \midrule
    Released                            & \wallBase{} & 52\% \\
    \rowcolor{black!8}
    + round-trip (ours)                 & \wallOurs{} & 10\% \\
    Corrected with the simulator        & \wallFixedAll{} & \phantom{0}5\% \\
    Corrected from the frames           & \wallFixedImg{} & \phantom{0}3\% \\
    \bottomrule
  \end{tabular}
\end{table}

\parahead{With the check, model error misleads the planner less.}
To see why the check works, we execute 50 candidates of an iteration and compare their predicted and true costs.
With CEM alone on Wall, on average 85\% of a difference in predicted cost between two candidates carries over to their true costs at iteration 10, but only 44\% at iteration 20.
The rest is model error, so CEM then selects mostly on it, which is how its plans turn adversarial.
With $r$ in the cost, this share stays at 80\% at iteration 20.

\begin{table}[t]
  \centering
  \caption{\textbf{Early stopping, the per-action target and more search stay below the check.}
  Success rate of the final plan without replanning.
  The budget is the number of candidates times the number of iterations, and $300\times30$ is the released one.
  Early stopping takes the iteration chosen on the separate problems, and the best iteration is the best one on the test problems.}
  \label{tab:analysis}
  \small
  \setlength{\tabcolsep}{4.2pt}
  \begin{tabular}{@{}llccc}
    \toprule
    Budget & Method & Wall & Cube & TwoRoom \\
    \midrule
                                   & CEM                         & \wallBase{} & \cubeBase{} & \twoBase{} \\
                                   & CEM with early stopping     & \wallEarly{} & \cubeEarly{} & \twoEarly{} \\
                                   & CEM at its best iteration   & \wallBasePeak{} & \cubeBasePeak{} & \twoBasePeak{} \\
                                   & + per-action at $\beta{=}0.5$ & 0.64 & 0.68 & 0.77 \\
                                   & + per-action at $\beta{=}1$   & 0.52 & 0.66 & 0.76 \\
    \rowcolor{black!8}
    \cellcolor{white}              & + round-trip at $\beta{=}0.5$ & 0.72 & \textbf{0.81} & 0.81 \\
    \rowcolor{black!8}
    \cellcolor{white}\multirow{-7}{*}{$300\times30$} & + round-trip at $\beta{=}1$ & \textbf{0.76} & 0.79 & \textbf{0.82} \\
    \midrule
                                   & CEM                         & 0.45 & 0.59 & 0.76 \\
    \rowcolor{black!8}
    \cellcolor{white}\multirow{-2}{*}{$900\times30$} & + round-trip (ours) & \textbf{0.78} & \textbf{0.78} & \textbf{0.85} \\
    \midrule
                                   & CEM                         & 0.28 & 0.64 & 0.75 \\
    \rowcolor{black!8}
    \cellcolor{white}\multirow{-2}{*}{$300\times60$} & + round-trip (ours) & \textbf{0.65} & \textbf{0.82} & \textbf{0.80} \\
    \bottomrule
  \end{tabular}
\end{table}

\parahead{The identifiable target matters more than the weight.}
At both weights in \Cref{tab:analysis}, the block-sum target is ahead of the per-action target on every task, as \cref{sec:certificate} predicts.
Its success changes by at most 0.04 between the two weights, while the per-action target gets worse with the larger one on every task.
On TwoRoom, the block-sum IDM reaches a validation $R^2$ of 0.96 (1 is a perfect prediction), while a per-action IDM reaches $-0.04$, no better than predicting the mean.
A per-action IDM with more than twenty times the parameters and three times the training overfits and raises the validation $R^2$ by at most 0.02.
ACID's flow-matching IDM~\cite{acid2026} takes a single Euler step, which returns the conditional mean that least squares also estimates.
No network can recover what the states do not determine.

\parahead{Other remedies stay below the check.}
On Wall and Cube, stopping a single search early cannot replace the check.
In \Cref{tab:analysis}, CEM stays below it both at the iteration chosen on the separate problems and at its best iteration on the test problems, while the check runs all 30 iterations.
A larger budget does not help either, as CEM gains nothing significant from three times as many candidates or twice as many iterations, while the check stays ahead at both budgets.
On Wall, an ensemble baseline in the spirit of model-based reinforcement learning~\cite{yu2020mopo,kidambi2020morel} penalizes disagreement among four retrained predictors and the released one, using the same normalization as the check.
It reaches 0.61, below the \wallOurs{} of the check ($p=0.002$).
The check thus needs no corrected model, more search or predictors.
The adversarial view points it to the plan's prediction, and identifiability to the block sum, read back by a small network.
\section{Conclusion}
\label{sec:conclusion}

Planning with a latent world model optimizes actions against the model's own prediction, much as an adversarial example optimizes an image against a classifier.
We showed that a longer search by CEM or MPPI can turn genuine plans into adversarial ones, and that the model cannot recognize them.
Our test measures how often this happens and tells which planners need repair.
Unlike an image, a plan can be checked against its own prediction.
Since two consecutive predicted states determine only the sum of the actions between them, our round-trip check reads this sum back with one small network and penalizes plans whose actions do not add up to it.
On three released world models, it raises the success of the CEM and MPPI planners that the test finds in need of repair, without retraining and at only \ovhWall\% to \ovhCube\% more planning time.
With the check, model error misleads the planner less, and neither early stopping nor a larger search budget matches it.
On Wall it raises the success of the final plan from \wallBase{} to \wallOurs{} without correcting the position labels that the adversarial plans exploit.

A normal validation loss and a high success with replanning can hide how often a single search ends in an adversarial plan, and the adversarial rate makes this visible.
Beyond latent world models, we expect the same behavior wherever a planner optimizes against a learned model, such as a learned dynamics or reward model.
We hope that this adversarial view opens new directions for how world models are trained, evaluated and used for planning.
\par
{
  \small
  \bibliographystyle{ieeenat_fullname}
  \bibliography{main}

\begin{thebibliography}{56}
\providecommand{\natexlab}[1]{#1}
\providecommand{\url}[1]{\texttt{#1}}
\expandafter\ifx\csname urlstyle\endcsname\relax
  \providecommand{\doi}[1]{doi: #1}\else
  \providecommand{\doi}{doi: \begingroup \urlstyle{rm}\Url}\fi

\bibitem[Agrawal et~al.(2016)Agrawal, Nair, Abbeel, Malik, and
  Levine]{agrawal2016poke}
Pulkit Agrawal, Ashvin Nair, Pieter Abbeel, Jitendra Malik, and Sergey Levine.
\newblock Learning to poke by poking: Experiential learning of intuitive
  physics.
\newblock In \emph{Advances in Neural Information Processing Systems
  (NeurIPS)}, 2016.

\bibitem[Agrawal et~al.(2026)Agrawal, Kalra, Parthasarathy, LeCun, Bounou,
  Izmailov, and Goldblum]{agrawal2025closing}
Rohun Agrawal, Nimit Kalra, Arjun Parthasarathy, Yann LeCun, Oumayma Bounou,
  Pavel Izmailov, and Micah Goldblum.
\newblock Closing the train-test gap in world models for gradient-based
  planning.
\newblock \emph{arXiv preprint arXiv:2512.09929}, 2026.

\bibitem[Alrasheed et~al.(2026)Alrasheed, Azam, and Akhtar]{planninglimits2026}
Ali Alrasheed, Basim Azam, and Naveed Akhtar.
\newblock The planning limits of latent world models.
\newblock \emph{arXiv preprint arXiv:2609.39235}, 2026.

\bibitem[Assran et~al.(2025)Assran, Bardes, Fan, Garrido, Howes, Komeili,
  Muckley, Rizvi, Roberts, Sinha, Zholus, Arnaud, Gejji, Martin, Hogan, Dugas,
  Bojanowski, Khalidov, Labatut, Massa, Szafraniec, Krishnakumar, Li, Ma,
  Chandar, Meier, LeCun, Rabbat, and Ballas]{assran2025vjepa2}
Mido Assran, Adrien Bardes, David Fan, Quentin Garrido, Russell Howes, Mojtaba
  Komeili, Matthew Muckley, Ammar Rizvi, Claire Roberts, Koustuv Sinha, Artem
  Zholus, Sergio Arnaud, Abha Gejji, Ada Martin, Francois~Robert Hogan, Daniel
  Dugas, Piotr Bojanowski, Vasil Khalidov, Patrick Labatut, Francisco Massa,
  Marc Szafraniec, Kapil Krishnakumar, Yong Li, Xiaodong Ma, Sarath Chandar,
  Franziska Meier, Yann LeCun, Michael Rabbat, and Nicolas Ballas.
\newblock {V-JEPA} 2: Self-supervised video models enable understanding,
  prediction and planning.
\newblock \emph{arXiv preprint arXiv:2506.09985}, 2025.

\bibitem[Athalye et~al.(2018)Athalye, Carlini, and
  Wagner]{athalye2018obfuscated}
Anish Athalye, Nicholas Carlini, and David Wagner.
\newblock Obfuscated gradients give a false sense of security: Circumventing
  defenses to adversarial examples.
\newblock In \emph{International Conference on Machine Learning (ICML)}, 2018.

\bibitem[Baker et~al.(2022)Baker, Akkaya, Zhokhov, Huizinga, Tang, Ecoffet,
  Houghton, Sampedro, and Clune]{baker2022vpt}
Bowen Baker, Ilge Akkaya, Peter Zhokhov, Joost Huizinga, Jie Tang, Adrien
  Ecoffet, Brandon Houghton, Raul Sampedro, and Jeff Clune.
\newblock Video pretraining ({VPT}): Learning to act by watching unlabeled
  online videos.
\newblock In \emph{Advances in Neural Information Processing Systems
  (NeurIPS)}, 2022.

\bibitem[Boney et~al.(2019)Boney, Di~Palo, Berglund, Ilin, Kannala, Rasmus, and
  Valpola]{boney2019regularizing}
Rinu Boney, Norman Di~Palo, Mathias Berglund, Alexander Ilin, Juho Kannala,
  Antti Rasmus, and Harri Valpola.
\newblock Regularizing trajectory optimization with denoising autoencoders.
\newblock In \emph{Advances in Neural Information Processing Systems
  (NeurIPS)}, 2019.

\bibitem[Carlini and Wagner(2017{\natexlab{a}})]{carlini2017detected}
Nicholas Carlini and David Wagner.
\newblock Adversarial examples are not easily detected: Bypassing ten detection
  methods.
\newblock In \emph{ACM Workshop on Artificial Intelligence and Security
  (AISec)}, 2017{\natexlab{a}}.

\bibitem[Carlini and Wagner(2017{\natexlab{b}})]{carlini2017towards}
Nicholas Carlini and David Wagner.
\newblock Towards evaluating the robustness of neural networks.
\newblock In \emph{IEEE Symposium on Security and Privacy (SP)},
  2017{\natexlab{b}}.

\bibitem[Chi et~al.(2023)Chi, Feng, Du, Xu, Cousineau, Burchfiel, and
  Song]{chi2023diffusion}
Cheng Chi, Siyuan Feng, Yilun Du, Zhenjia Xu, Eric Cousineau, Benjamin
  Burchfiel, and Shuran Song.
\newblock Diffusion policy: Visuomotor policy learning via action diffusion.
\newblock In \emph{Robotics: Science and Systems (RSS)}, 2023.

\bibitem[Christiano et~al.(2016)Christiano, Shah, Mordatch, Schneider,
  Blackwell, Tobin, Abbeel, and Zaremba]{christiano2016transfer}
Paul Christiano, Zain Shah, Igor Mordatch, Jonas Schneider, Trevor Blackwell,
  Joshua Tobin, Pieter Abbeel, and Wojciech Zaremba.
\newblock Transfer from simulation to real world through learning deep inverse
  dynamics model.
\newblock \emph{arXiv preprint arXiv:1610.03518}, 2016.

\bibitem[Chua et~al.(2018)Chua, Calandra, McAllister, and Levine]{chua2018pets}
Kurtland Chua, Roberto Calandra, Rowan McAllister, and Sergey Levine.
\newblock Deep reinforcement learning in a handful of trials using
  probabilistic dynamics models.
\newblock In \emph{Advances in Neural Information Processing Systems
  (NeurIPS)}, 2018.

\bibitem[Deng et~al.(2009)Deng, Dong, Socher, Li, Li, and
  Fei-Fei]{deng2009imagenet}
Jia Deng, Wei Dong, Richard Socher, Li-Jia Li, Kai Li, and Li Fei-Fei.
\newblock {ImageNet}: A large-scale hierarchical image database.
\newblock In \emph{IEEE Conference on Computer Vision and Pattern Recognition
  (CVPR)}, 2009.

\bibitem[Dosovitskiy et~al.(2021)Dosovitskiy, Beyer, Kolesnikov, Weissenborn,
  Zhai, Unterthiner, Dehghani, Minderer, Heigold, Gelly, Uszkoreit, and
  Houlsby]{dosovitskiy2021vit}
Alexey Dosovitskiy, Lucas Beyer, Alexander Kolesnikov, Dirk Weissenborn,
  Xiaohua Zhai, Thomas Unterthiner, Mostafa Dehghani, Matthias Minderer, Georg
  Heigold, Sylvain Gelly, Jakob Uszkoreit, and Neil Houlsby.
\newblock An image is worth 16x16 words: Transformers for image recognition at
  scale.
\newblock In \emph{International Conference on Learning Representations
  (ICLR)}, 2021.

\bibitem[Du et~al.(2023)Du, Yang, Dai, Dai, Nachum, Tenenbaum, Schuurmans, and
  Abbeel]{du2023unipi}
Yilun Du, Mengjiao Yang, Bo Dai, Hanjun Dai, Ofir Nachum, Joshua~B. Tenenbaum,
  Dale Schuurmans, and Pieter Abbeel.
\newblock Learning universal policies via text-guided video generation.
\newblock In \emph{Advances in Neural Information Processing Systems
  (NeurIPS)}, 2023.

\bibitem[Eykholt et~al.(2018)Eykholt, Evtimov, Fernandes, Li, Rahmati, Xiao,
  Prakash, Kohno, and Song]{eykholt2018robust}
Kevin Eykholt, Ivan Evtimov, Earlence Fernandes, Bo Li, Amir Rahmati, Chaowei
  Xiao, Atul Prakash, Tadayoshi Kohno, and Dawn Song.
\newblock Robust physical-world attacks on deep learning visual classification.
\newblock In \emph{IEEE Conference on Computer Vision and Pattern Recognition
  (CVPR)}, 2018.

\bibitem[Fu et~al.(2021)Fu, Kumar, Nachum, Tucker, and Levine]{fu2020d4rl}
Justin Fu, Aviral Kumar, Ofir Nachum, George Tucker, and Sergey Levine.
\newblock {D4RL}: Datasets for deep data-driven reinforcement learning.
\newblock \emph{arXiv preprint arXiv:2004.07219}, 2021.

\bibitem[Gao et~al.(2023)Gao, Schulman, and Hilton]{gao2023scaling}
Leo Gao, John Schulman, and Jacob Hilton.
\newblock Scaling laws for reward model overoptimization.
\newblock In \emph{International Conference on Machine Learning (ICML)}, 2023.

\bibitem[Goodfellow et~al.(2015)Goodfellow, Shlens, and
  Szegedy]{goodfellow2015explaining}
Ian~J. Goodfellow, Jonathon Shlens, and Christian Szegedy.
\newblock Explaining and harnessing adversarial examples.
\newblock In \emph{International Conference on Learning Representations
  (ICLR)}, 2015.

\bibitem[Ha and Schmidhuber(2018)]{ha2018recurrent}
David Ha and J{\"u}rgen Schmidhuber.
\newblock Recurrent world models facilitate policy evolution.
\newblock In \emph{Advances in Neural Information Processing Systems
  (NeurIPS)}, 2018.

\bibitem[Henaff et~al.(2018)Henaff, Whitney, and LeCun]{henaff2017model}
Mikael Henaff, William~F. Whitney, and Yann LeCun.
\newblock Model-based planning with discrete and continuous actions.
\newblock \emph{arXiv preprint arXiv:1705.07177}, 2018.

\bibitem[Hendrycks and Gimpel(2023)]{hendrycks2016gelu}
Dan Hendrycks and Kevin Gimpel.
\newblock Gaussian error linear units ({GELUs}).
\newblock \emph{arXiv preprint arXiv:1606.08415}, 2023.

\bibitem[Hu et~al.(2026)Hu, Zheng, Bo, Wang, Dubost, Mottini, Liu, Srinivasan,
  Zhong, Qian, Gao, and Cui]{scale2026}
Jiaming Hu, Yan Zheng, Shi Bo, Tian Wang, Florian Dubost, Alejandro Mottini,
  Junze Liu, Arvind Srinivasan, Kai Zhong, Kun Qian, Sharon Gao, and Qingjun
  Cui.
\newblock Reperesentation geometry matters for planning with {JEPA} world
  models.
\newblock \emph{arXiv preprint arXiv:2608.16287}, 2026.

\bibitem[Ioffe and Szegedy(2015)]{ioffe2015batchnorm}
Sergey Ioffe and Christian Szegedy.
\newblock Batch normalization: Accelerating deep network training by reducing
  internal covariate shift.
\newblock In \emph{International Conference on Machine Learning (ICML)}, 2015.

\bibitem[Janner et~al.(2021)Janner, Li, and Levine]{janner2021trajectory}
Michael Janner, Qiyang Li, and Sergey Levine.
\newblock Offline reinforcement learning as one big sequence modeling problem.
\newblock In \emph{Advances in Neural Information Processing Systems
  (NeurIPS)}, 2021.

\bibitem[Janner et~al.(2022)Janner, Du, Tenenbaum, and
  Levine]{janner2022diffuser}
Michael Janner, Yilun Du, Joshua Tenenbaum, and Sergey Levine.
\newblock Planning with diffusion for flexible behavior synthesis.
\newblock In \emph{International Conference on Machine Learning (ICML)}, 2022.

\bibitem[Jing and Boularias(2025)]{wmvae2025}
Eric Jing and Abdeslam Boularias.
\newblock Bounding distributional shifts in world modeling through novelty
  detection.
\newblock In \emph{IEEE/RSJ International Conference on Intelligent Robots and
  Systems (IROS)}, 2025.

\bibitem[Jordan and Rumelhart(1992)]{jordan1992forward}
Michael~I. Jordan and David~E. Rumelhart.
\newblock Forward models: Supervised learning with a distal teacher.
\newblock \emph{Cognitive Science}, 16\penalty0 (3):\penalty0 307--354, 1992.

\bibitem[Kidambi et~al.(2020)Kidambi, Rajeswaran, Netrapalli, and
  Joachims]{kidambi2020morel}
Rahul Kidambi, Aravind Rajeswaran, Praneeth Netrapalli, and Thorsten Joachims.
\newblock {MOReL}: Model-based offline reinforcement learning.
\newblock In \emph{Advances in Neural Information Processing Systems
  (NeurIPS)}, 2020.

\bibitem[Kurakin et~al.(2017)Kurakin, Goodfellow, and
  Bengio]{kurakin2017physical}
Alexey Kurakin, Ian Goodfellow, and Samy Bengio.
\newblock Adversarial examples in the physical world.
\newblock In \emph{International Conference on Learning Representations (ICLR)
  Workshop}, 2017.

\bibitem[Lee et~al.(2018)Lee, Lee, Lee, and Shin]{lee2018simple}
Kimin Lee, Kibok Lee, Honglak Lee, and Jinwoo Shin.
\newblock A simple unified framework for detecting out-of-distribution samples
  and adversarial attacks.
\newblock In \emph{Advances in Neural Information Processing Systems
  (NeurIPS)}, 2018.

\bibitem[Liu et~al.(2026)Liu, Wu, Wen, You, Zhang, Hao, Lin, and Su]{djepa2026}
Shuaijun Liu, Chengyu Wu, Qifu Wen, Feiyang You, Chenglong Zhang, Shuyang Hao,
  Xi Lin, and Ningxin Su.
\newblock {D-JEPA}: A decision-aligned latent world model.
\newblock \emph{arXiv preprint arXiv:2609.24749}, 2026.

\bibitem[Madry et~al.(2018)Madry, Makelov, Schmidt, Tsipras, and
  Vladu]{madry2018towards}
Aleksander Madry, Aleksandar Makelov, Ludwig Schmidt, Dimitris Tsipras, and
  Adrian Vladu.
\newblock Towards deep learning models resistant to adversarial attacks.
\newblock In \emph{International Conference on Learning Representations
  (ICLR)}, 2018.

\bibitem[Maes et~al.(2026)Maes, Le~Lidec, Scieur, LeCun, and
  Balestriero]{lewm2026}
Lucas Maes, Quentin Le~Lidec, Damien Scieur, Yann LeCun, and Randall
  Balestriero.
\newblock {LeWorldModel}: Stable end-to-end joint-embedding predictive
  architecture from pixels.
\newblock In \emph{Advances in Neural Information Processing Systems
  (NeurIPS)}, 2026.

\bibitem[Moosavi-Dezfooli et~al.(2016)Moosavi-Dezfooli, Fawzi, and
  Frossard]{moosavi2016deepfool}
Seyed-Mohsen Moosavi-Dezfooli, Alhussein Fawzi, and Pascal Frossard.
\newblock {DeepFool}: A simple and accurate method to fool deep neural
  networks.
\newblock In \emph{IEEE Conference on Computer Vision and Pattern Recognition
  (CVPR)}, 2016.

\bibitem[Nguyen et~al.(2015)Nguyen, Yosinski, and Clune]{nguyen2015fooled}
Anh Nguyen, Jason Yosinski, and Jeff Clune.
\newblock Deep neural networks are easily fooled: High confidence predictions
  for unrecognizable images.
\newblock In \emph{IEEE Conference on Computer Vision and Pattern Recognition
  (CVPR)}, 2015.

\bibitem[Oquab et~al.(2024)Oquab, Darcet, Moutakanni, Vo, Szafraniec, Khalidov,
  Fernandez, Haziza, Massa, El-Nouby, Assran, Ballas, Galuba, Howes, Huang, Li,
  Misra, Rabbat, Sharma, Synnaeve, Xu, Jegou, Mairal, Labatut, Joulin, and
  Bojanowski]{oquab2024dinov2}
Maxime Oquab, Timoth{\'e}e Darcet, Th{\'e}o Moutakanni, Huy~V. Vo, Marc
  Szafraniec, Vasil Khalidov, Pierre Fernandez, Daniel Haziza, Francisco Massa,
  Alaaeldin El-Nouby, Mido Assran, Nicolas Ballas, Wojciech Galuba, Russell
  Howes, Po-Yao Huang, Shang-Wen Li, Ishan Misra, Michael Rabbat, Vasu Sharma,
  Gabriel Synnaeve, Hu Xu, Herve Jegou, Julien Mairal, Patrick Labatut, Armand
  Joulin, and Piotr Bojanowski.
\newblock {DINO}v2: Learning robust visual features without supervision.
\newblock \emph{Transactions on Machine Learning Research}, 2024.

\bibitem[Park et~al.(2025)Park, Frans, Eysenbach, and Levine]{park2025ogbench}
Seohong Park, Kevin Frans, Benjamin Eysenbach, and Sergey Levine.
\newblock {OGBench}: Benchmarking offline goal-conditioned {RL}.
\newblock In \emph{International Conference on Learning Representations
  (ICLR)}, 2025.

\bibitem[Pathak et~al.(2017)Pathak, Agrawal, Efros, and
  Darrell]{pathak2017curiosity}
Deepak Pathak, Pulkit Agrawal, Alexei~A. Efros, and Trevor Darrell.
\newblock Curiosity-driven exploration by self-supervised prediction.
\newblock In \emph{International Conference on Machine Learning (ICML)}, 2017.

\bibitem[Pham and Bera(2026)]{leap2026}
Phu Pham and Aniket Bera.
\newblock Latent energy action planning with world models.
\newblock \emph{arXiv preprint arXiv:2609.03294}, 2026.

\bibitem[Ruan et~al.(2026)Ruan, Hsiao, Lo, and Shuai]{ruan2026future}
Bo-Kai Ruan, Teng-Fang Hsiao, Ling Lo, and Hong-Han Shuai.
\newblock Is the future compatible? {D}iagnosing dynamic consistency in world
  action models.
\newblock \emph{arXiv preprint arXiv:2605.07514}, 2026.

\bibitem[Rubinstein and Kroese(2004)]{rubinstein1999cem}
Reuven~Y. Rubinstein and Dirk~P. Kroese.
\newblock \emph{The Cross-Entropy Method: A Unified Approach to Combinatorial
  Optimization, Monte-Carlo Simulation and Machine Learning}.
\newblock Springer, 2004.

\bibitem[Seo et~al.(2026)Seo, Kim, and Kwak]{acid2026}
Gawon Seo, Dongwon Kim, and Suha Kwak.
\newblock {ACID}: Action consistency via inverse dynamics for planning with
  world models.
\newblock \emph{arXiv preprint arXiv:2607.02403}, 2026.

\bibitem[Smith and Winkler(2006)]{smith2006optimizer}
James~E. Smith and Robert~L. Winkler.
\newblock The optimizer's curse: Skepticism and postdecision surprise in
  decision analysis.
\newblock \emph{Management Science}, 52\penalty0 (3):\penalty0 311--322, 2006.

\bibitem[Sobal et~al.(2025)Sobal, Zhang, Cho, Balestriero, Rudner, and
  LeCun]{sobal2025pldm}
Uladzislau Sobal, Wancong Zhang, Kyunghyun Cho, Randall Balestriero, Tim G.~J.
  Rudner, and Yann LeCun.
\newblock Learning from reward-free offline data: A case for planning with
  latent dynamics models.
\newblock In \emph{Advances in Neural Information Processing Systems
  (NeurIPS)}, 2025.

\bibitem[Sun et~al.(2022)Sun, Ming, Zhu, and Li]{sun2022knn}
Yiyou Sun, Yifei Ming, Xiaojin Zhu, and Yixuan Li.
\newblock Out-of-distribution detection with deep nearest neighbors.
\newblock In \emph{International Conference on Machine Learning (ICML)}, 2022.

\bibitem[Szegedy et~al.(2014)Szegedy, Zaremba, Sutskever, Bruna, Erhan,
  Goodfellow, and Fergus]{szegedy2014intriguing}
Christian Szegedy, Wojciech Zaremba, Ilya Sutskever, Joan Bruna, Dumitru Erhan,
  Ian Goodfellow, and Rob Fergus.
\newblock Intriguing properties of neural networks.
\newblock In \emph{International Conference on Learning Representations
  (ICLR)}, 2014.

\bibitem[Tassa et~al.(2018)Tassa, Doron, Muldal, Erez, Li, de~Las~Casas,
  Budden, Abdolmaleki, Merel, Lefrancq, Lillicrap, and
  Riedmiller]{tassa2018dmcontrol}
Yuval Tassa, Yotam Doron, Alistair Muldal, Tom Erez, Yazhe Li, Diego de
  Las~Casas, David Budden, Abbas Abdolmaleki, Josh Merel, Andrew Lefrancq,
  Timothy Lillicrap, and Martin Riedmiller.
\newblock {DeepMind} control suite.
\newblock \emph{arXiv preprint arXiv:1801.00690}, 2018.

\bibitem[Terver et~al.(2026)Terver, Yang, Ponce, Bardes, and
  LeCun]{terver2025jepawms}
Basile Terver, Tsung-Yen Yang, Jean Ponce, Adrien Bardes, and Yann LeCun.
\newblock What drives success in physical planning with joint-embedding
  predictive world models?
\newblock \emph{Transactions on Machine Learning Research (TMLR)}, 2026.

\bibitem[Torabi et~al.(2018)Torabi, Warnell, and Stone]{torabi2018bco}
Faraz Torabi, Garrett Warnell, and Peter Stone.
\newblock Behavioral cloning from observation.
\newblock In \emph{International Joint Conference on Artificial Intelligence
  (IJCAI)}, 2018.

\bibitem[Trabucco et~al.(2021)Trabucco, Kumar, Geng, and
  Levine]{trabucco2021coms}
Brandon Trabucco, Aviral Kumar, Xinyang Geng, and Sergey Levine.
\newblock Conservative objective models for effective offline model-based
  optimization.
\newblock In \emph{International Conference on Machine Learning (ICML)}, 2021.

\bibitem[Wang et~al.(2022)Wang, Li, Feng, and Zhang]{wang2022vim}
Haoqi Wang, Zhizhong Li, Litong Feng, and Wayne Zhang.
\newblock {ViM}: Out-of-distribution with virtual-logit matching.
\newblock In \emph{IEEE/CVF Conference on Computer Vision and Pattern
  Recognition (CVPR)}, 2022.

\bibitem[Williams et~al.(2017)Williams, Wagener, Goldfain, Drews, Rehg, Boots,
  and Theodorou]{williams2017mppi}
Grady Williams, Nolan Wagener, Brian Goldfain, Paul Drews, James~M. Rehg, Byron
  Boots, and Evangelos~A. Theodorou.
\newblock Information theoretic {MPC} for model-based reinforcement learning.
\newblock In \emph{IEEE International Conference on Robotics and Automation
  (ICRA)}, 2017.

\bibitem[You et~al.(2026)You, Zhang, Ran, Yang, Zhang, Xue, Song, Tian, and
  Guo]{you2026control}
Hanzhe You, Yonggang Zhang, Maohao Ran, Zhiqin Yang, Zhenyuan Zhang, Wei Xue,
  Jun Song, Xinmei Tian, and Yike Guo.
\newblock A control theory of predictability in latent world models.
\newblock \emph{arXiv preprint arXiv:2607.10362}, 2026.

\bibitem[Yu et~al.(2020)Yu, Thomas, Yu, Ermon, Zou, Levine, Finn, and
  Ma]{yu2020mopo}
Tianhe Yu, Garrett Thomas, Lantao Yu, Stefano Ermon, James Zou, Sergey Levine,
  Chelsea Finn, and Tengyu Ma.
\newblock {MOPO}: Model-based offline policy optimization.
\newblock In \emph{Advances in Neural Information Processing Systems
  (NeurIPS)}, 2020.

\bibitem[Zhou et~al.(2025)Zhou, Pan, LeCun, and Pinto]{zhou2025dinowm}
Gaoyue Zhou, Hengkai Pan, Yann LeCun, and Lerrel Pinto.
\newblock {DINO-WM}: World models on pre-trained visual features enable
  zero-shot planning.
\newblock In \emph{International Conference on Machine Learning (ICML)}, 2025.

\end{thebibliography}
}

\end{document}